\pdfoutput=1

\documentclass[11pt]{article}

\usepackage[preprint]{acl}

\usepackage{times}
\usepackage{latexsym}

\usepackage[T1]{fontenc}

\usepackage[utf8]{inputenc}

\usepackage{microtype}

\usepackage{inconsolata}

\usepackage{graphicx}

\usepackage{amsmath}
\usepackage{amssymb}
\usepackage{mathtools}
\usepackage{amsthm}

\newcommand{\yhat}{\hat{y}}

\usepackage[utf8]{inputenc} 
\usepackage[T1]{fontenc}    
\usepackage{hyperref}       
\usepackage{url}            
\usepackage{booktabs}       
\usepackage{amsfonts}       
\usepackage{nicefrac}       
\usepackage{microtype}      
\usepackage{xcolor}         
\usepackage{graphicx}
\usepackage{multirow, multicol, makecell}  
\usepackage{pifont}
\usepackage{color, colortbl}
\usepackage[most]{tcolorbox}
\usepackage{array}
\usepackage{color}

\usepackage{threeparttable}
\usepackage{subfigure}
\usepackage{dsfont}
\usepackage{enumerate}
\usepackage{multirow}
\usepackage{longtable}
\usepackage{wrapfig} 
\usepackage{newtxtext}

\definecolor{darkgreen}{rgb}{0.0, 0.5, 0.0}

\title{CoT-UQ: Improving Response-wise Uncertainty Quantification in LLMs with Chain-of-Thought}

\author{Boxuan Zhang \\
  Purdue University \\
  \texttt{zhan5479@purdue.edu} \\\And
  Ruqi Zhang \\
  Purdue University \\
  \texttt{ruqiz@purdue.edu} \\}

\begin{document}
\maketitle
\begin{abstract}
Large language models (LLMs) excel in many tasks but struggle to accurately quantify uncertainty in their generated responses. This limitation makes it challenging to detect misinformation and ensure reliable decision-making.
Existing uncertainty quantification (UQ) methods for LLMs are primarily prompt-wise rather than response-wise, often requiring multiple response samples, which incurs high computational costs. Moreover, LLMs have been shown to be overconfident, particularly when using reasoning steps to derive their answers.
In this work, we propose \emph{CoT-UQ}, a response-wise UQ framework that integrates LLMs’ inherent reasoning capabilities through Chain-of-Thought (CoT) into the UQ process. CoT-UQ captures critical information during inference by extracting keywords from each reasoning step and assessing their importance to the final answer. This key reasoning information is then aggregated to produce a final uncertainty estimate.
We conduct extensive experiments based on Llama Family with model sizes varying from 8B to 13B across logical and mathematical reasoning tasks.
Experimental results demonstrate that CoT-UQ significantly outperforms existing UQ methods, achieving an average improvement of 5.9\% AUROC compared to current UQ methods.
The code is available at: \url{https://github.com/ZBox1005/CoT-UQ}
\end{abstract}

\section{Introduction}
\label{sec: intro}
Large language models (LLMs) have demonstrated groundbreaking capabilities across a variety of applications \citep{ouyang2022training, chowdhery2023palm, openai2024b, guo2025deepseek}. Particularly, prompting techniques like Chain-of-Thought (CoT) \citep{wei2022chain} have significantly enhanced LLMs reasoning capabilities, ranging from multi-round conversation \citep{long2023large, chen2023chatcot}, logical reasoning \citep{creswell2022selection, duan2024gtbench} and mathematical reasoning \citep{yao2024tree, shao2024deepseekmath}. However, LLMs often unpredictably hallucinate \citep{manakul2023selfcheckgpt}, i.e., making plausible but incorrect statements \citep{ji2023survey}, limiting their deployment in safety-critical applications \citep{clusmann2023future}.

\begin{figure}[t!]
  \centering
  \includegraphics[width=0.48\textwidth]{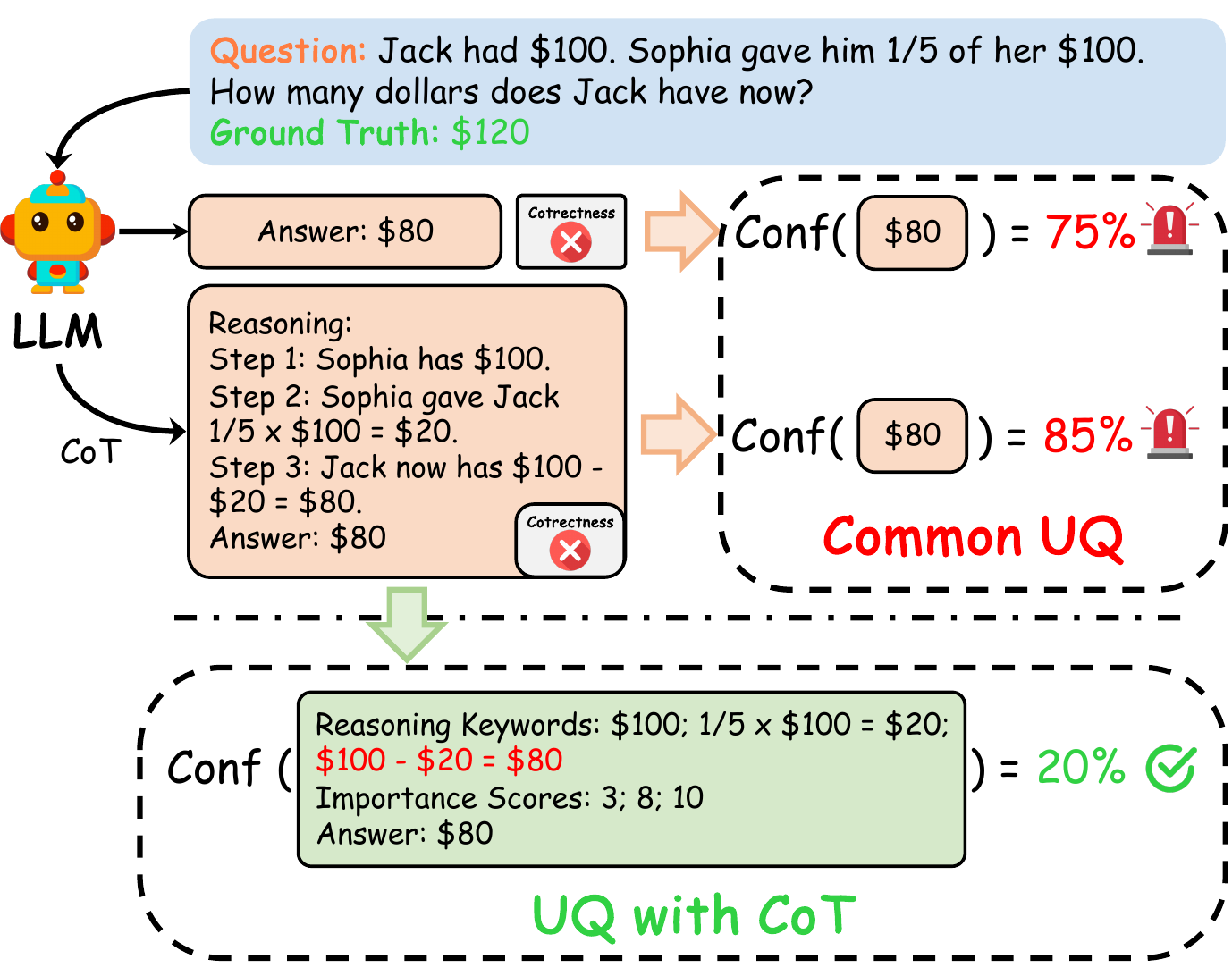}
  \vspace{-6mm}
  \caption{Comparison of existing UQ strategies with ours. Directly estimating the uncertainty of a generated incorrect answer leads to overconfidence, which is exacerbated by using CoT to derive the answer. 
  We tackle this challenge by integrating CoT into the UQ process with keywords extraction and importance scores.
  } 
  \vspace{-2mm}
  \label{fig:motivation}
\end{figure}

To improve the reliability of LLMs, uncertainty quantification (UQ) has emerged as a key strategy for determining when humans can trust LLM-generated outputs. However, existing UQ methods for LLMs are primarily prompt-wise \citep{malinin2021uncertainty, kuhn2023semantic, ling2024uncertainty}. That is, uncertainty is calculated at the prompt level rather than for each individual response. These methods require multiple response samples per prompt, leading to high computational costs and inference time. Besides, some studies \citep{kadavath2022language, miao2023selfcheck} propose leveraging an LLM’s own ability to evaluate the uncertainty of its responses without relying on external knowledge. However, these approaches suffer from overconfidence issues, particularly when reasoning steps, such as Chain-of-Thought (CoT), are used before deriving the final answer \citep{fu2025multiple}. Overconfidence has been attributed to the model’s inherent bias toward trusting its own outputs \citep{mielke2022reducing, lin2022teaching}.

To enable response-wise reliable UQ, we propose utilizing LLMs' powerful reasoning capabilities in the UQ process. Our motivation comes from an intuitive insight: providing access to the reasoning path allows the model to use additional context for confidence calibration, leading to a more informed assessment of the final answer’s uncertainty.
As illustrated in Figure \ref{fig:motivation}, while LLMs tend to be overconfident in their generated answers, and using CoT can further amplify this issue, we hypothesize that incorporating key reasoning information into the UQ process can effectively mitigate inflated confidence scores, resulting in better-calibrated uncertainty estimates. This naturally raises a critical research question:
\textit{How can LLMs’ reasoning steps be utilized to improve uncertainty estimation?}

\begin{figure*}[t!]
  \centering
  \includegraphics[width=0.96\textwidth]{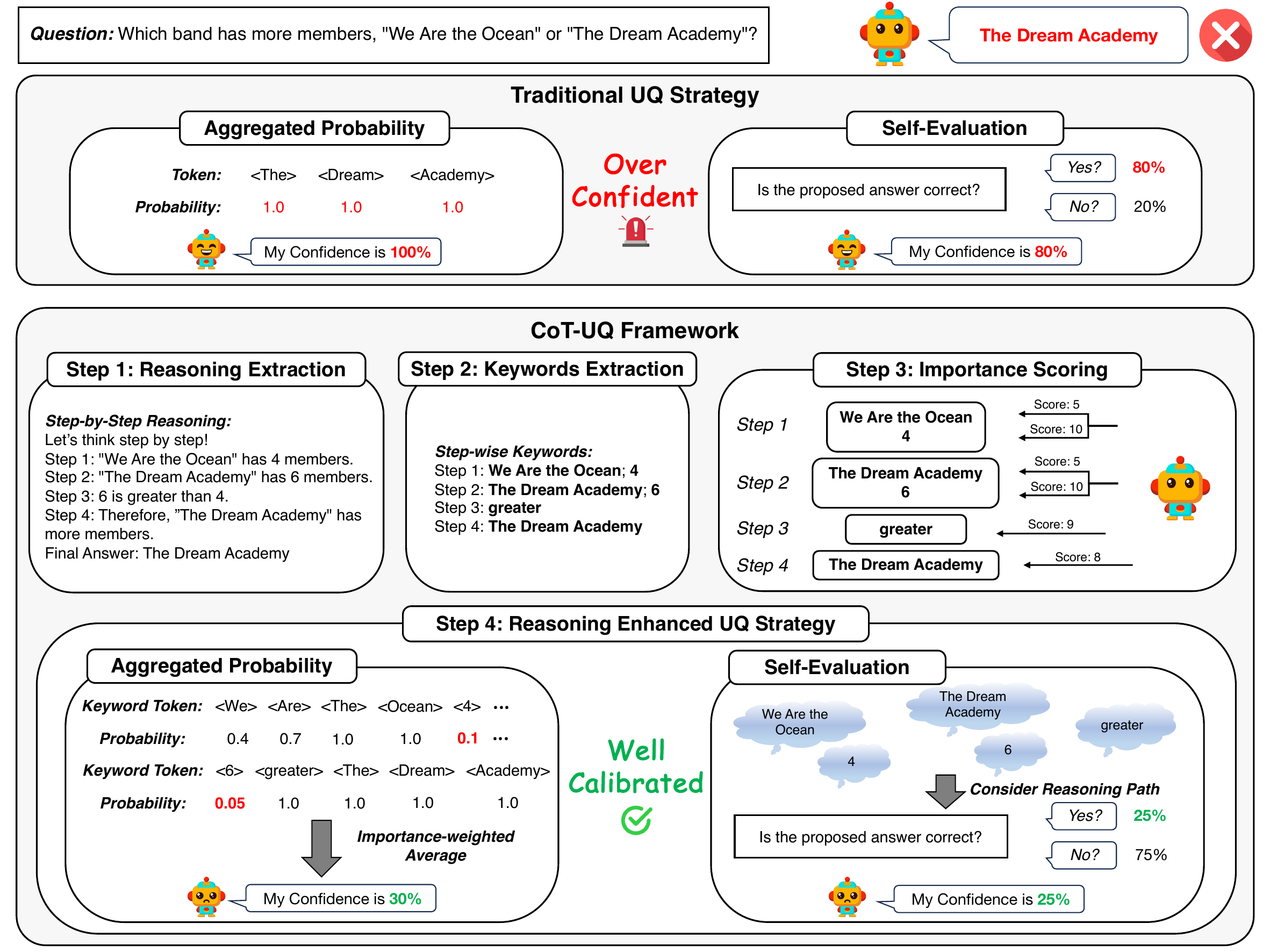}
  \caption{Illustration of our framework, CoT-UQ. Given a question and an incorrect response generated by an LLM, the \textit{top} of the figure shows two common UQ strategies, which suffer from overconfidence issues. The \textit{bottom} shows the four-step process of CoT-UQ: performing the reasoning process, extracting step-wise keywords, scoring the importance of keywords relative to the final answer, and leveraging reasoning information to enhance common UQ strategies. CoT-UQ leads to a better-calibrated response-wise uncertainty estimate.} 
  \label{fig:framework}
  \vspace{-2mm}
\end{figure*}

To answer this question, we propose a new framework, namely, \textbf{C}hain-\textbf{o}f-\textbf{T}hought enhanced \textbf{U}ncertainty \textbf{Q}uantification (CoT-UQ). 
At a high level, CoT-UQ provides response-wise uncertainty by integrating reasoning steps from CoT into the UQ process.
Specifically, as illustrated in Figure~\ref{fig:framework}, CoT-UQ leverages the LLM’s reasoning process to extract keywords from each inference step and assess their importance in determining the final answer. By incorporating this critical information, CoT-UQ achieves better-calibrated uncertainty estimation, either by aggregating token probabilities of extracted keywords or by integrating the reasoning path into the self-evaluation process.

We conducted extensive experiments to verify the effectiveness of our proposed framework. 
Under extensive evaluations, CoT-UQ achieves superior performance compared with different baselines, which reveals that LLMs have the potential to use their own reasoning to better express the trustworthiness of their generations. We also conduct a range of ablation studies and provide in-depth discussions via case studies. Our contributions can be summarized as the following:
\begin{itemize}
    \item Conceptually, we introduce a novel perspective to quantify uncertainty for LLMs by leveraging LLM's internal reasoning knowledge.
    \item Technically, we propose a new UQ framework, namely, \textbf{C}hain-\textbf{o}f-\textbf{T}hought enhanced \textbf{U}ncertainty \textbf{Q}uantification (CoT-UQ), which integrates the reasoning knowledge into the UQ process through extracting keywords and evaluating corresponding importance scores. 
    \item Empirically, we conduct extensive experiments on the Llama family across five datasets in two tasks and show that CoT-UQ achieves an average AUROC improvement of approximately 5.9\% compared to baselines, which verifies the effectiveness of our method.
\end{itemize}

\section{Related Works}

\subsection{Uncertainty Quantification in LLMs}
Prior efforts to quantify uncertainty and confidence in LLMs can be categorized into four main approaches. The first approach is to derive calibrated confidence by examining agreement across multiple sampled responses \citep{malinin2021uncertainty, kuhn2023semantic, manakul2023selfcheckgpt, tian2023fine}. However, as \citet{qiu2024semantic} recently pointed out, these methods primarily quantify prompt-wise rather than response-wise uncertainty. While \citet{qiu2024semantic} provides a method for response-wise uncertainty, it still relies on generating multiple response samples, making it computationally inefficient. The second approach is to leverage LLM's own ability to evaluate the confidence of its responses, often through self-probing techniques \citep{kadavath2022language, tian2023just, xiong2023can}. The third approach is to aggregate token probabilities of its generated response, which includes adopting traditional UQ methods \citep{xiao2022uncertainty, ye2024benchmarking} and assigning importance weights to tokens \citep{duan2024shifting, bakman2024mars}. However, the above two approaches still suffer from overconfidence due to the model’s inherent bias to trust its own outputs \citep{mielke2022reducing, lin2022teaching}. The fourth approach is to fine-tune the original LLM to calibrate its confidence \citep{lin2022teaching, kapoor2024large}. However, the model-specific tuning has limited their applications to new scenarios. In contrast to these four approaches, the proposed CoT-UQ is a response-wise UQ method that does not require response sampling or model-specific tuning. Instead, it leverages the LLM’s inherent reasoning abilities to calibrate uncertainty/confidence scores, making it readily generalizable to new tasks and models. 

\subsection{Chain of Thought Reasoning in LLMs}
To equip LLMs with capabilities to solve more complex and reasoning tasks, \citet{wei2022chain} extended in-context learning by introducing the concept of Chain of Thought (CoT) through a step-by-step reasoning process. \citet{kojima2022large} found that simply adding a leading sentence “Let’s think step by step” to a cue allowed LLMs to perform zero-shot logical reasoning without any additional human prompts \citep{chu2023survey}. Subsequently, CoT-SC~\citep{wang2022self} introduces a self-consistency strategy to replace the greedy decoding strategy. \citet{feng2024towards} further reveals the underlying mechanisms behind CoT through a theoretical perspective. \citet{liu2024era} refines CoT by capturing relationships between entities to aid LLMs in understanding context. Although these studies highlight the importance of CoT in enhancing LLMs’ reasoning abilities in various situations, a recent study \citep{fu2025multiple} observes that CoT can exacerbate the overconfident issues in LLMs when only measuring the final answer. To the best of our knowledge, CoT-UQ is the first approach to integrate reasoning knowledge into the UQ process for LLMs.

\section{Preliminaries}
\label{sec: prelimi}
In this section, we briefly introduce the preliminaries of response-wise uncertainty quantification~(UQ) in LLMs, including problem settings and two popular UQ strategies, namely aggregated probabilities and self-evaluation. Further details on these two strategies can be found in Appendix \ref{app:details:baseline}.

\paragraph{Problem Setups.}
Given an LLM $M$, an input prompt $p$, and the output sequence $\yhat = [y_1, y_2, ..., y_L]$, where $L$ is the number of tokens generated by LLM, the task is to obtain a confidence score for users representing the probability that $\yhat$ is correct. 
In the following, we illustrate the existing two paradigms for response-wise uncertainty quantification and analyze their limitations.

\paragraph{Aggregated Probabilities (\textit{AP}).}
Previous works \citep{kadavath2022language, huang2023look, varshney2023stitch} based on aggregated probabilities typically aggregate output token probabilities of the generated text tokens $\yhat = [y_1, y_2, ..., y_L]$ to measure the LLM's confidence for each response. For the type of aggregation techniques, we consider several methods following \citet{orgad2024llms}, including the mean and the minimum of these values. Formally, given an aggregation function $\textit{Aggr}(\cdot)$, the confidence score $c$ can be abstracted as,
\begin{equation}
\label{eq:1}
    c = \mathop{\textit{Aggr}}\limits_{i=1}^{N}(\mathbb{P}(y_i | p, y_1, ..., y_{i-1})).
\end{equation}

\paragraph{Self-Evaluation (\textit{SE}).}
Self-evaluation strategies~\citep{kadavath2022language, xiong2023can} usually contain a two-stage process to elicit the confidence score from LLMs: 1) Using an input comprising of the question $q$ combined with the prompt $p$ to generate the text response $\yhat$. 2) Combining $q$ and $\yhat$ through a well-designed prompt $p^{t}$ to instruct LLM to self-evaluate the correctness of $\yhat$. Among them, one representative baseline is P(True)~\citep{kadavath2022language}. P(True) is straightforward yet effective by directly asking LLM whether the predicted $\yhat$ is true or false to $q$ via $p^{t}$ and using the probability of “True” as confidence $c$, which is defined as follows,
\begin{equation}
\label{eq:2}
    c = \mathbb{P}(o = \textit{True}), \quad\text{where}\ o = M(p^{t}(q, \yhat)).
\end{equation}

Although previous UQ methods using AP and SE have shown promising results, they often suffer from overconfidence, particularly when using Chain-of-Thought reasoning for complex tasks.  
In this work, we explore how to leverage the intrinsic reasoning capabilities of LLMs to mitigate overconfidence. Specifically, we propose extracting step-wise keywords from the model’s inference process and integrating this knowledge into AP and SE strategies to better assess the trustworthiness of LLM-generated outputs.

\section{Methodology}
\label{sec: method}
In this section, we introduce our new framework, i.e., Chain-of-Thought enhanced Uncertainty Quantification (CoT-UQ), as illustrated in Figure \ref{fig:framework}.
Unlike common UQ strategies that directly assess the confidence of the generated answer, CoT-UQ is a two-stage paradigm containing four steps during inference. The first three steps focus on refining multi-step reasoning by extracting keywords and their corresponding importance scores to the final answer (Section \ref{sec: method: stage1}). The fourth step illustrates how to integrate this crucial reasoning information into the two common UQ strategies~(Section \ref{sec: method: stage2}). 

\subsection{Stage 1: LLM Inference Refining}
\label{sec: method: stage1}
\paragraph{Step 1: Reasoning Extraction.}
We first instruct LLM to derive the reasoning for each response. Before inference, we add the step-wise Chain-of-Thought (CoT) prefix for prompting, i.e., "\textit{Let’s think step by step. Step 1:}". This ensures the model’s inference results are structured into multiple reasoning steps, with each step explicitly starting with "\textit{Step i:}". Upon completion of inference, we obtain a response $\hat{y}$ for the question $q$, which contains a step-by-step reasoning $s_{1 \sim k} = s_1, . . . , s_k$ and a final answer $a$ labeled with ``\textit{Final Answer:}''.

\paragraph{Step 2: Keywords Extraction.}
After obtaining the step-by-step inference $s_{1 \sim k}$ for each question-answer pair $(q, a)$, we choose to extract keywords from each step. 
Prior works generally consider the sum or average of all generated tokens \citep{slobodkin2023curious} to aggregate token-level uncertainty. 
However, these strategies potentially introduce redundant tokens, which can significantly compromise the accuracy of uncertainty scores.\citep{gupta2024language}.
This motivates us to consider tokens from keywords, which better represent the most meaningful part of a reasoning step. Specifically, we request the LLM itself to complete the extraction, as LLMs have demonstrated information extraction capability \citep{ashok2023promptner, orgad2024llms}. Formally, we extract $n_i \ge 0$ keywords from each reasoning step $s_i\in s_{1\sim k}$ (noted $n_i=0$ means no effective keywords in a specific step, we explain this situation in Appendix \ref{app:imple}). The keywords set $\mathcal{K}$ extracted from all steps can be formulated as, 
\begin{equation}
\label{eq:3}
    \mathcal{K} = \bigcup_{i=1}^k \{w_j^i\}_{j=1}^{n_i},
\end{equation}
where $\{w_j^i\}_{j=1}^{n_i}$ are the keywords extracted from step $i$.

\paragraph{Step 3: Importance Scoring.}
Relying on the self-evaluation \citep{ren2023self} capability, we instruct the LLM to rate the importance of the keywords in deriving the final answer in a few-shot in-context learning setup. We provide the context (question $q$, multi-step reasoning $s_{1 \sim k}$, and final answer $a$) combined with the keywords $\mathcal{K}$ extracted in Step 2 to the LLM. Each keyword will be scored by the LLM, ranging from 1 to 10, where 1 denotes the least critical and 10 is the most. For keywords that are more critical in the inference, i.e., require exact reasoning or imply vital elements to the final answer, we assign a higher score to this keyword. For instance, as shown in Figure \ref{fig:framework}, keywords that reveal the specific number of members in each band will get a higher importance score, even to 10, as they explicitly require reasoning and are crucially contributing to the final answer. After integrating with the corresponding importance score $t$, keywords set $\mathcal{K}$ can be updated as,
\begin{equation}
\label{eq:4}
    \mathcal{K} = \bigcup_{i=1}^k \{(w_j^i, t_j^i)\}_{j=1}^{n_i}.
\end{equation}
In the first three steps, we deconstructed the redundant reasoning steps into keywords containing the most meaningful and critical information and evaluated their respective importance towards reaching the final answer, formalizing them as a keywords set containing dualist formulation. In the next step, we will use these keywords to help enhance existing uncertainty quantification strategies.

\subsection{Stage 2 / Step 4: Reasoning Enhanced Uncertainty Quantification Strategy}
\label{sec: method: stage2}
Given the extracted reasoning path $s_{1 \sim k}$ and keywords set $\mathcal{K}$ defined in Section \ref{sec: method: stage1}, we aim to elicit the confidence $c$ and mitigate overconfidence issues generally exist in common UQ strategies, \textit{aggregated probabilities} (\textit{AP}) and \textit{self-evaluation} (\textit{SE}). We propose an integration method for each strategy to utilize the information provided by $s_{1 \sim k}$ and $\mathcal{K}$.

\paragraph{Reasoning Enhanced \textit{Aggregated Probabilities} (\textit{AP}) Strategy.}
Compared to directly combining the token probabilities from output tokens, we choose to aggregate those from the extracted keywords to integrate the inference knowledge into the UQ process. Since the keywords are generally short in token length, and for the sake of comparison, we use the same aggregation techniques $\mathop{\textit{Aggr}}(\cdot)$ introduced in Section \ref{sec: prelimi} to aggregate the token probabilities of a single keyword into its prediction probability. Formally, given a single keyword $w$ (text) with its corresponding token sequence $\hat{w} = [w_1, w_2, ..., w_l]$ of length $l$, the probability of the keyword $w$ can be formulated as,
\begin{equation}
\label{eq:6}
    p(\hat{w}) = \mathop{\textit{Aggr}} \limits_{m=1}^l (\mathbb{P}(w_m \mid p, w_1, \dots, w_{m-1})).
\end{equation}
To consider their contributions to the final confidence, we propose to average the probabilities of keywords weighted by their importance scores. This ensures that more significant keywords have a greater influence on the final confidence estimation. The final confidence estimation can be formalized as follows:
\begin{equation}
\begin{aligned}
\label{eq:5}
    c = \frac{\sum_{i=1}^{k}\sum_{j=1}^{n_i}t_j^i \cdot p(\hat{w_j^i}) }{\sum_{i=1}^{k}\sum_{j=1}^{n_i}t_j^i}.
\end{aligned}
\end{equation}

\begin{table*}[h!]
    \centering
    \resizebox{0.99\linewidth}{!}{
        \begin{tabular}{c|c|l|cc|ccc}
            \toprule
             \multirow{2}*{\textbf{Model}} & \multirow{2}*{\textbf{Strategy}} & \multirow{2}*{\textbf{Method}} & \multicolumn{2}{c}{\textbf{Logical Reasoning}} & \multicolumn{3}{c}{\textbf{Mathematical Reasoning}} \\ 
             \cmidrule{4-5}\cmidrule{6-8}
             & & & HotpotQA & 2WikiMHQA & GSM8K & SVAMP & ASDiv \\
            \midrule
            \multirow{10}*{\textbf{Llama 3.1-8B}} & \multirow{6}*{\textit{AP}} & Probas-mean & 53.73 & 56.80 & 53.17 & 53.94 & 58.34 \\
            & &  \textbf{w/ CoT-UQ} & \textbf{62.01} & \textbf{65.22} & \textbf{63.64} & \textbf{59.83} & \textbf{64.52} \\
            \cmidrule{3-8}
            & & Probas-min & 58.34 & 56.81 & 54.95 & 54.79 & 58.69 \\
            & &  \textbf{w/ CoT-UQ} & \textbf{64.37} & \textbf{70.02} & \textbf{63.09} & \textbf{60.49} & \textbf{64.84}  \\
            \cmidrule{3-8}
            & & {\small TOKEN}\textit{SAR} & 53.57 & 56.92 & 54.46 & 55.01  & 58.71 \\
            & &  \textbf{w/ CoT-UQ} & \textbf{61.07} & \textbf{65.38} & \textbf{65.10} & \textbf{62.11} & \textbf{66.91} \\
            \cmidrule{2-8}
            & \multirow{4}*{\textit{SE}} & P(True) & 62.39 & 53.56 & 48.15 & 51.58 & 47.23 \\
            & &  \textbf{w/ CoT-UQ} & \textbf{63.10} & \textbf{57.77} & \textbf{52.60} & \textbf{60.00} & \textbf{53.20} \\
            \cmidrule{3-8}
            & & Self-Probing & 54.33 & 56.39 & 49.24 & 51.63 & 50.86 \\
            & &  \textbf{w/ CoT-UQ} & \textbf{57.20}  & \textbf{58.38} & \textbf{51.89} & \textbf{54.26} & \textbf{53.79} \\
            \midrule
            \multirow{10}*{\textbf{Llama 2-13B}} & \multirow{6}*{\textit{AP}} & Probas-mean & 56.27 & 51.54 & 53.96 & 54.48 & 57.73 \\
            & &  \textbf{w/ CoT-UQ} & \textbf{66.56} & \textbf{63.29} & \textbf{58.54} & \textbf{57.37} & \textbf{59.44} \\
            \cmidrule{3-8}
            & & Probas-min & 56.51 & 51.28 & 53.84 & 55.09 & 57.70 \\
            & &  \textbf{w/ CoT-UQ} & \textbf{67.19} & \textbf{68.10} & \textbf{58.63} & \textbf{58.51} & \textbf{60.74} \\
            \cmidrule{3-8}
            & & {\small TOKEN}\textit{SAR} & 57.33 & 51.08 & 54.82 & 55.06 & 58.37 \\
            & &  \textbf{w/ CoT-UQ} & \textbf{66.29} & \textbf{64.03} & \textbf{59.61} & \textbf{58.41} & \textbf{61.23} \\
            \cmidrule{2-8}
            & \multirow{4}*{\textit{SE}} & P(True) & 51.13 & 47.52 & 46.06 & 46.36 & 48.02 \\
            & &  \textbf{w/ CoT-UQ} & \textbf{57.10} & \textbf{53.52} & \textbf{52.59} & \textbf{56.87} & \textbf{56.10} \\
            \cmidrule{3-8}
            & & Self-Probing & 60.63 & 59.81 & 52.72 & 47.27 & 52.35 \\
            & &  \textbf{w/ CoT-UQ} & \textbf{64.03} & \textbf{62.63} & \textbf{55.14} & \textbf{50.53} & \textbf{57.58} \\
            \bottomrule
        \end{tabular}
    }
    \caption{AUROC ($\uparrow$) comparison of Llama 3.1-8B and Llama 2-13B on various benchmarks for logical and mathematical reasoning tasks, where \textit{AP} indicates \textit{aggregated probabilities} and \textit{SE} denotes \textit{self-evaluation}. For the chosen type of \textit{SE} strategy, we propose {\small KEY}\textit{Keywords} for both two \textit{SE} strategies in logical reasoning tasks, {\small ALL}\textit{Steps} for P(True) and {\small KEY}\textit{Step} for Self-Probing in mathematical reasoning tasks (See Section \ref{sec:exp:main} for analysis).}
    \vspace{-2mm}
    \label{tab:main-results}
\end{table*}

\paragraph{Reasoning Enhanced \textit{Self-Evaluation} (\textit{SE}) Strategy.} 
We provide \textit{four} approaches to instruct the LLM to consider the reasoning information during the self-evaluation of uncertainty. 
The first two methods, namely, {\small ALL}\textit{Steps} and {\small ALL}\textit{Keywords}, directly add the extracted reasoning steps $s_{1 \sim k}$ or keywords set $\mathcal{K}$ to the self-evaluation process.
To highlight the role of relevant importance of extracted keywords, we further introduce {\small KEY}\textit{Step} and {\small KEY}\textit{Keywords} strategies. {\small KEY}\textit{Step} proposes to consider the most important step, where the importance is calculated from the average of the importance scores from each reasoning step. {\small KEY}\textit{Step} can be abstracted as follows,
\begin{equation}
\label{eq:7}    
s^* = \arg\max_{1 \le i \le k}
\Biggl(
\frac{1}{n_i}\sum_{j=1}^{n_i} t_j^i
\Biggr).
\end{equation}
Meanwhile, the goal of {\small KEY}\textit{Keywords} is to exclude redundant keywords and shortlist the most critical ones based on their importance. We formulate it as,
\begin{equation}
\label{eq:8}
    \mathcal{K}^* = \bigcup_{i=1}^{k} \left\{ (w_j^i, t_j^i) \mid t_j^i \ge \tau \right\}_{j=1}^{n_i},
\end{equation}
where $\tau$ is a threshold to filter the sub-critical keywords during self-evaluation. We discuss the sensitivity to this hyper-parameter in Section \ref{sec:exp:abla}.

We include the above information in the self-evaluation prompt $p^t$ through an additional instruction, termed \textit{Considering <reasoning type> as additional information} (See Appendix \ref{app:imple:prompt} for the concrete realization). Formally, given a type of reasoning knowledge $z \in \bigl\{s_{1 \sim k},\, s^*,\, \mathcal{K},\, \mathcal{K}^*\bigr\}$, the self-evaluation process can be updated by the refined prompt $p^r$ as,
\begin{equation}
\begin{aligned}
\label{eq:9}
    c = \mathbb{P}(o = \textit{True}), \ \text{where}\ o = M\bigl(p^{r}(q, \hat{y}, z)\bigr).
\end{aligned}
\end{equation}

\section{Experiments}
\label{sec: exp}

\subsection{Experimental Setups}
\label{sec:exp:steup}

\paragraph{Datasets and Models.}
We consider the following reasoning scenarios: logical reasoning and mathematical reasoning, where existing UQ strategies suffer from overconfidence issues. 
Specifically, for logical reasoning, we use the HotpotQA dataset \citep{yang2018hotpotqa} and the 2WikiMultiHopQA dataset \citep{ho2020constructing}; for mathematical reasoning, we use the GSM8K dataset \citep{cobbe2021training}, the SVAMP dataset \citep{patel2021nlp}, and the ASDiv dataset \citep{miao2021diverse}. 
For models, we use Llama 2-13B and Llama 3.1-8B \citep{touvron2023llama}. 
Details of dataset statistics are provided in Appendix \ref{app:details:data}, while the implementation details including LLMs’ hyper-parameters, prompts, and computational costs are provided in Appendix \ref{app:imple}.

\begin{figure*}[t!]
  \centering
  \includegraphics[width=0.98\textwidth]{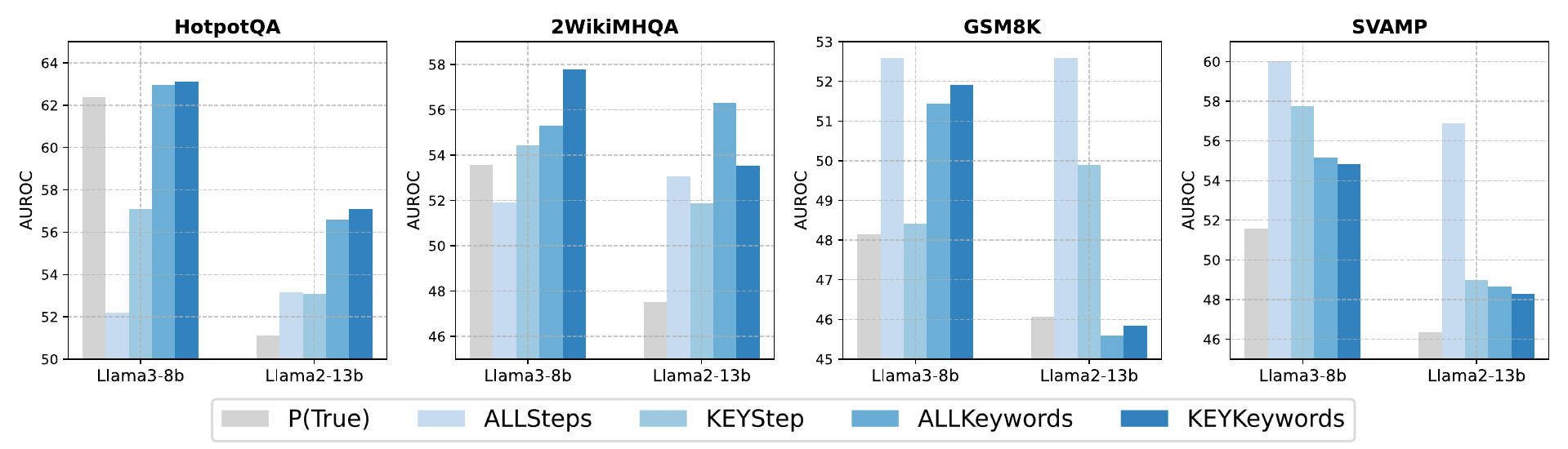}
  \vspace{-1mm}
  \caption{Comparison of different implementations of CoT-UQ for P(True). {\small KEY}\textit{Keywords} is more effective for logical reasoning tasks (HotpotQA and 2WikiMultiHopQA), whereas {\small ALL}\textit{Steps} works better for mathematical reasoning tasks (GSM8K and SVAMP).}
  \label{fig:exp:ptrue}
  \vspace{-1mm}
\end{figure*}

\begin{figure*}[t!]
  \centering
  \includegraphics[width=0.98\textwidth]{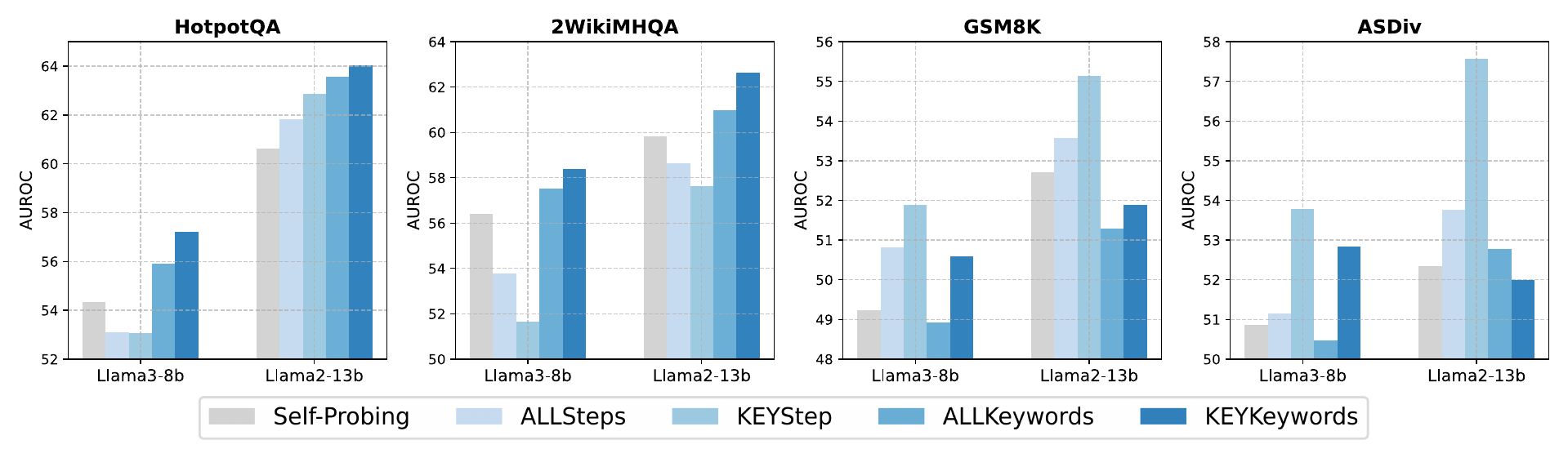}
  \vspace{-1mm}
  \caption{Comparison of different implementations of CoT-UQ based on Self-Probing. {\small KEY}\textit{Keywords} consistently achieves better performance on logical reasoning tasks (HotpotQA and 2WikiMultiHopQA), while {\small KEY}\textit{Step} performs better in mathematical reasoning (GSM8K and ASDiv).}
  \label{fig:exp:probing}
  \vspace{-1mm}
\end{figure*}

\paragraph{Evaluation Metric.}
Following the common evaluation approach in \citet{kuhn2023semantic}, we use the Area Under the Receiver Operating Characteristic curve (AUROC) \citep{davis2006relationship} to evaluate the performance of UQ methods, which measures the likelihood that a positive sample will receive a higher discriminating score than a negative sample \citep{fawcett2006introduction}. 
A higher AUROC score indicates better performance, while a score of 0.5 implies random guessing.

\paragraph{Baseline Methods.}
We compare the proposed framework with various competitive uncertainty quantification baseline methods. 
As mentioned in Section \ref{sec: prelimi}, common response-wise UQ strategies include aggregated probabilities (\textit{AP}) and self-evaluation (\textit{SE}). For \textit{AP}, inspired by \citep{kadavath2022language, guerreiro2022looking}, 
we first investigate the most common aggregation techniques \textbf{Probas-mean} and \textbf{Probas-min}. We also consider the Toekn-level Shifting Attention to Relevance (\textbf{{\small TOKEN}\textit{SAR}}) \citep{duan2024shifting} that evaluates the relevance of each token in the final answer and assigns higher weights to more relevant tokens.
For \textit{SE}, we consider \textbf{P(True)} \citep{kadavath2022language} and \textbf{Self-Probing} \citep{xiong2023can} that directly asks LLM to evaluate the correctness of their generation via prompting. Details on the implementations of baseline methods are listed in Appendix \ref{app:details:baseline}.

\begin{figure*}[t]
  \centering
  \subfigure[Aggregated Probabilities]{
      \centering
      \includegraphics[width=0.46\textwidth]{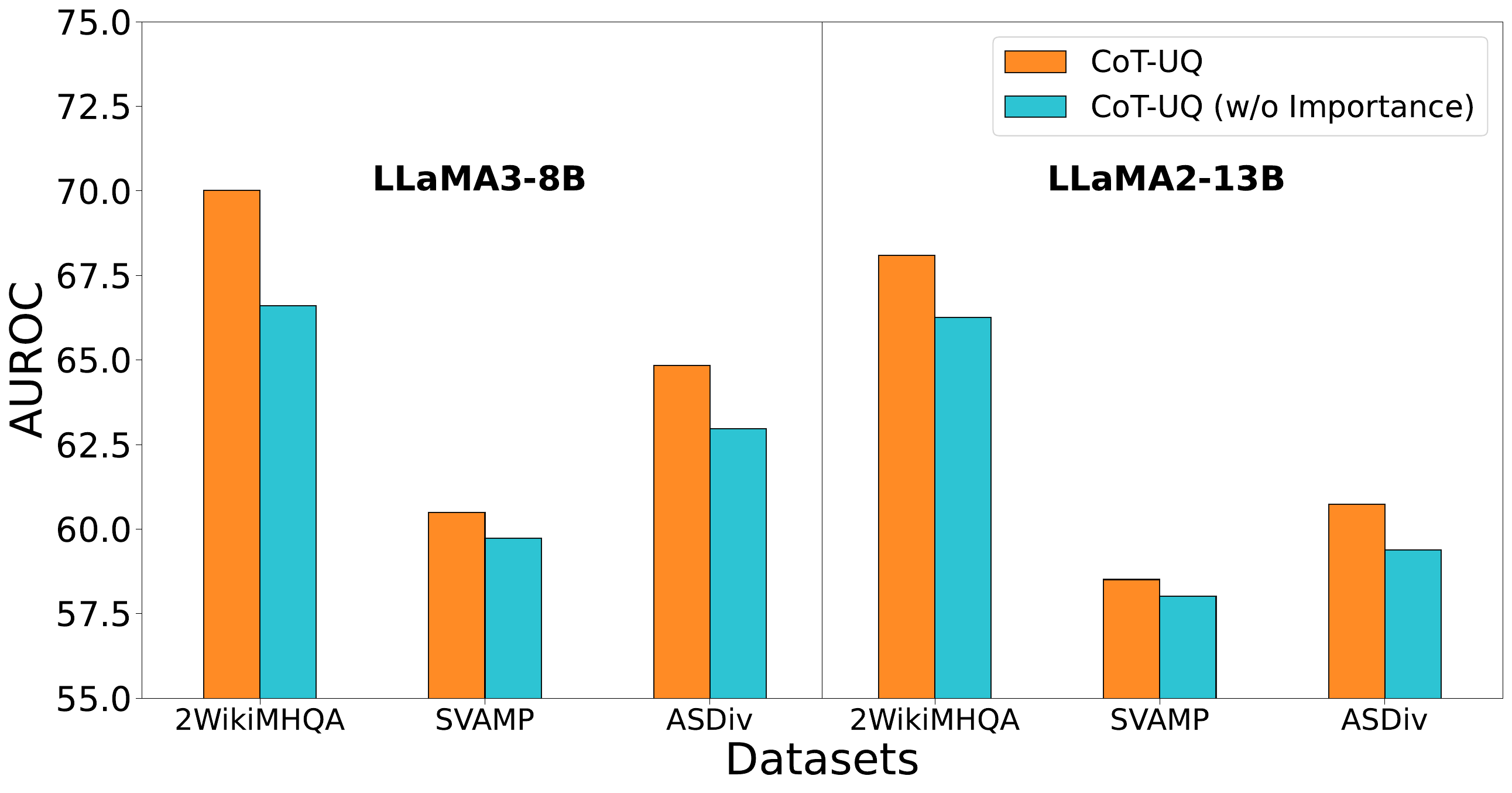}
  }
  \subfigure[Self-Evaluation]{
      \centering
      \includegraphics[width=0.46\textwidth]{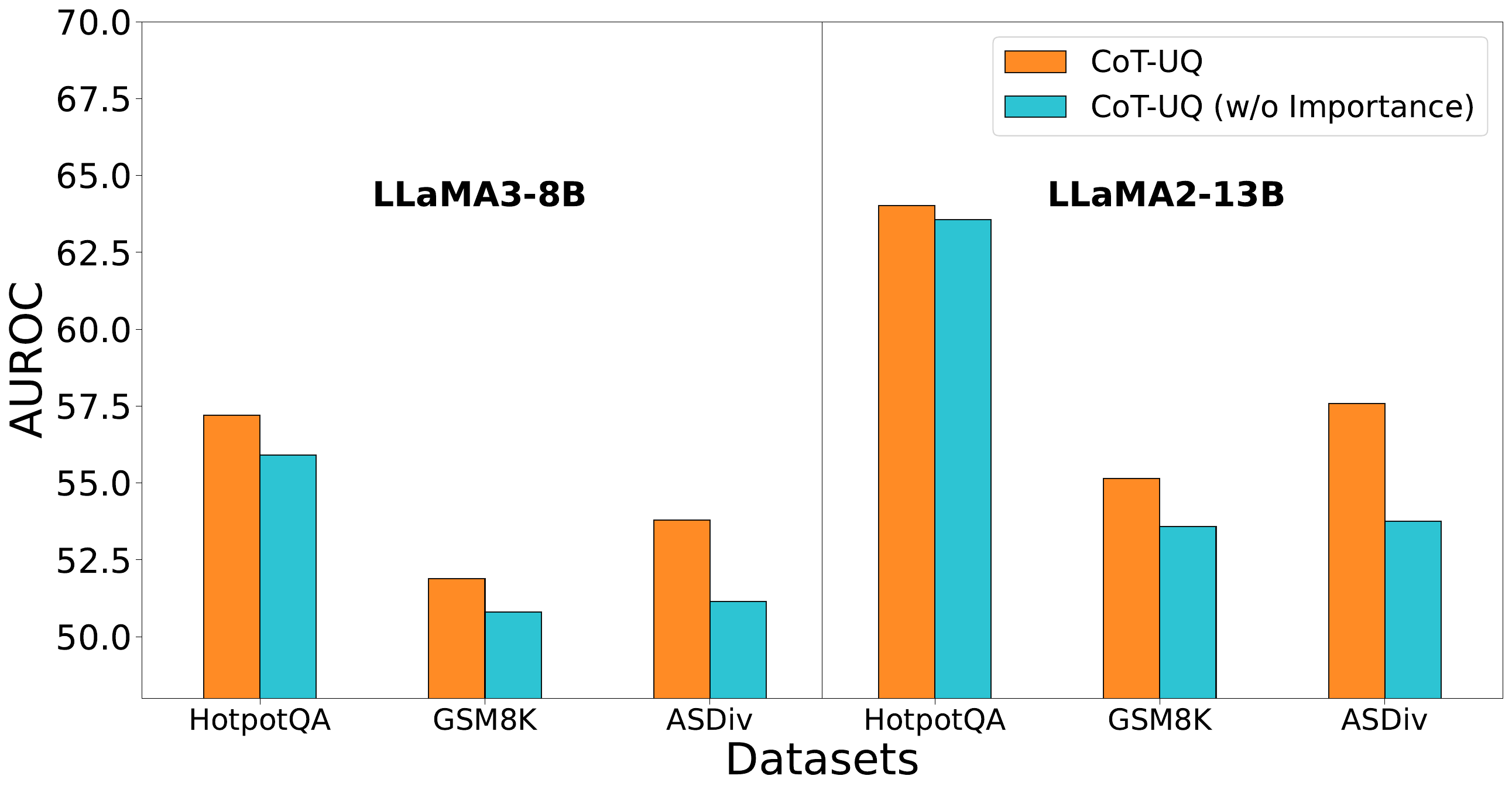}
  }
  \vspace{-2mm}
  \caption{Effect of importance scoring in CoT-UQ. (a) AUROC comparison of aggregated probability (\textbf{Probas-min}).(b) AUROC comparison of self-evaluation (\textbf{Self-Probing}).}
  \label{fig:importance}
  \vspace{-2mm}
\end{figure*}

\subsection{Main Results}
\label{sec:exp:main}
The overall comparison results are presented in Table \ref{tab:main-results}. We evaluate various baseline methods by comparing their performance with and without CoT-UQ.  
As shown in Table \ref{tab:main-results}, CoT-UQ consistently improves UQ performance across all tasks and datasets. This demonstrates that incorporating reasoning into uncertainty quantification enables LLMs to provide more calibrated assessments of the trustworthiness of their generated outputs. 
In general, CoT-UQ achieves greater improvements when applied to \textit{AP} strategies compared to \textit{SE} strategies, particularly for \textbf{Probas-min}, where it increases AUROC by up to \textbf{16.8\%}.

\paragraph{Aggregated Probabilities (\textit{AP}).}
When comparing CoT-UQ with three \textit{AP} strategies, our framework significantly outperforms them on the two logical reasoning datasets, HotpotQA and 2WikiMultiHopQA, achieving an average AUROC improvement of \textbf{+10.3\%} across both models.
Similarly, CoT-UQ improves performance on all three mathematical reasoning benchmarks.
It is worth noting that the recent work {\small TOKEN}\textit{SAR} gets limited improvements or even worse performance compared to Probas-mean on reasoning tasks, which may be caused by the limited length of the generated final answer in these tasks, especially in mathematical reasoning. CoT-UQ improves {\small TOKEN}\textit{SAR} significantly by aggregating keyword token probabilities, mitigating the impact of response length on uncertainty estimation.

\paragraph{Self-Evaluation (\textit{SE}).}
For the comparison with \textit{SE} strategies, we suggest different implementations of CoT-UQ across different reasoning tasks and \textit{SE} strategies. Specifically, we propose {\small KEY}\textit{Keywords} for both two \textit{SE} strategies in logical reasoning tasks, {\small ALL}\textit{Steps} for P(True) strategy and {\small KEY}\textit{Step} for Self-Probing strategy in mathematical reasoning tasks. 
These suggestions are based on the observation that keywords extracted from mathematical reasoning tend to be overly simplistic (e.g., single digits) and lack informative content. As a result, step-level strategies, which retain rich contextual information, are more suitable. In contrast, keywords extracted from logical reasoning generally retain meaningful and logical information, while complete reasoning steps may introduce redundant content that harms the model's judgment.

We report the results of recommended realizations in Table \ref{tab:main-results} and a detailed explanation and analysis for the above suggestions in Section \ref{sec:exp:abla}. The results of these five datasets demonstrate an average of \textbf{+4.4\%} improvement, highlighting the effectiveness of CoT-UQ compared to standard \textit{SE} baselines. 
Notably, CoT-UQ applied to \textit{SE} strategies shows a greater performance improvement in mathematical reasoning tasks (+5.3\%) compared to logical reasoning tasks (+3.5\%). This suggests that incorporating reasoning in UQ allows the model to identify critical errors in the thought process during self-evaluation, especially in mathematical problems where a single misstep can lead to incorrect conclusions. We provide detailed evidence and analysis via various case studies in Appendix \ref{app:add:case}.

\subsection{Ablation and Future Discussions}
\label{sec:exp:abla}
In this section, we provide a thorough understanding of our CoT-UQ. Additional results and discussions (e.g., comprehensive validation across model families and metric, sensitivity to hyper-parameters, importance scores visualization, and several case studies) can be found in Appendix \ref{app:add}.

\paragraph{Effect of Different Implementation of CoT-UQ in \textit{SE} Strategies.}
We observe an interesting trend: the transition from step-level implementations ({\small ALL}\textit{Steps} and {\small KEY}\textit{Step}) to keywords-level implementations ({\small ALL}\textit{Keywords} and {\small KEY}\textit{Keywords}) shows opposite trends in logical and mathematical reasoning tasks, as demonstrated in Figure \ref{fig:exp:ptrue} and Figure \ref{fig:exp:probing}. Specifically, {\small ALL}\textit{Keywords} and {\small KEY}\textit{Keywords} perform significantly better than step-level techniques on HotpotQA and 2WikiMHQA datasets. This suggests that step-level information usually contains redundant words in logical reasoning tasks, and the keywords effectively filter out irrelevant information for uncertainty estimation. Conversely, {\small ALL}\textit{Steps} consistently performs well when applied to P(True) on GSM8K and SVAMP datasets, and {\small KEY}\textit{Step} performs similarly when applied to Self-Probing. However, keywords-level methods show suboptimal performance, or even worse than the standard SE on mathematical datasets. This may be attributed to the necessity of including sufficient context when assessing mathematical answers, rather than relying on a few scattered keywords.

\paragraph{Effect of Importance Scoring.}
The effects of the importance scoring step are summarized in Figure \ref{fig:importance}, where we employ Probas-min and Self-Probing to represent the \textit{AP} and \textit{SE} strategies, respectively.
For \textit{AP}, w/o importance indicates directly calculating the mean of probabilities from keywords.
For \textit{SE},  we use the specific implementations as above suggested for different reasoning tasks. For instance, we adopt the {\small KEY}\textit{Keywords} for HotpotQA, and the {\small KEY}\textit{Step} for mathematical reasoning benchmarks.
Here, w/o importance in \textit{SE} demotes the corresponding realizations start with {\small ALL}, i.e, {\small ALL}\textit{Keywords} and {\small ALL}\textit{Steps}.
The results highlight the necessity of evaluating the respective importance metric for each keyword in our method.

\section{Conclusion}
In this paper, we introduce \textbf{C}hain-\textbf{o}f-\textbf{T}hought enhanced \textbf{U}ncertainty \textbf{Q}uantification (CoT-UQ), the first approach that leveraging LLM's internal knowledge through CoT for uncertainty quantification. CoT-UQ consistently and significantly boosts the performance of current aggregated probabilities and self-evaluation strategies by using crucial information from the reasoning path.
We conduct extensive experiments to validate the effectiveness of our framework and provide detailed ablation studies and discussions.

\section*{Ethics Statement}
The datasets we used are sourced from the current public datasets. The prompts we used do not collect or use personal information or information from other individuals. Furthermore, they do not contain any sensitive words or oppose any individual or group.
CoT-UQ has the potential to impact the credibility and reliability of LLMs, particularly in the context of reducing misinformation. LLMs have the potential to generate highly plausible but false information. Uncertainty quantification techniques can help distinguish between accurate and misleading outputs. Successfully addressing this issue can help prevent the spread of misinformation and mitigate its potential societal impact.

\section*{Limitations}
Our methods require access to token logits. Although commercial LLM providers widely support token logits, this still might restrict the potential application of our methods in black-box scenarios.
In addition, our framework is limited to the closed-ended question-answering domain, where a question has an objective ground-truth answer(s) so that we can justify the correctness of generated answer. Extensive analysis of CoT-UQ on open-ended question-answering tasks is beyond the scope of the current study and is left as future work.

\bibliography{acl_latex}

\clearpage
\appendix

\section{Details about Considered Datasets and Baselines}
\label{app:details}

\subsection{Datasets.}
\label{app:details:data}
\paragraph{Detailed introductions.}
We outline here all five datasets belonging to two reasoning domains that we investigate in our work. For each dataset that has been divided into training and test sets, we used samples from its test set unless otherwise instructed.
\begin{itemize}
    \item \textbf{HotpotQA} \citep{yang2018hotpotqa}: a dataset designed for diverse multi-hop question answering. Each entry includes Wikipedia documents that help answering the questions. We use the setting without context, where questions are asked directly without additional context.
    \item \textbf{2WikiMHQA} \citep{ho2020constructing}: a dataset uses both structured and unstructured data. The dataset also introduces the evidence information containing a reasoning path for multi-hop questions. There are four types of questions: \textit{comparison}, \textit{inference}, \textit{compositional}, and \textit{bridge-comparison}.
    In this paper, we use the \textit{inference} type of questions in all experiments.
    \item \textbf{GSM8K} \citep{cobbe2021training}: a dataset of 8.5K high quality linguistically diverse grade school math word problems created by human problem writers. These problems take between 2 and 8 steps to solve, and solutions primarily involve performing a sequence of elementary calculations using basic arithmetic operations to reach the final answer.
    \item \textbf{SVAMP}\citep{patel2021nlp}: a challenge dataset for elementary-level Math Word Problems (MWP). An MWP consists of a short Natural Language narrative that describes a state of the world and poses a question about some unknown quantities.
    \item \textbf{ASDiv} \citep{miao2021diverse}: a diverse (in terms of both language patterns and problem types) English math word problem (MWP) corpus for evaluating the capability of various MWP solvers. Each MWP is annotated with its problem type and grade level (for indicating the level of difficulty).
\end{itemize}

\paragraph{Dataset Statistics}
Table \ref{app:tab:dataset} provides detailed information about the data included in the experiment, with a minimum of 1000 samples and a total of 14563 samples taken.
\begin{table}[h]
    \centering
    \resizebox{0.99\linewidth}{!}{\begin{tabular}{llll}
        \toprule
        \textbf{Dataset}& \textbf{Num.} & \textbf{Length} & \textbf{Domain} \\
        \midrule
         HotpotQA & 8447 & 23.2 & Logical Reasoning \\
         2WikiMHQA & 1548 & 14.5 & Logical Reasoning   \\
         \midrule
         GSM8K & 1,319 & 58.9 & Mathematical Reasoning  \\
         SVAMP & 1000 & 39.4 & Mathematical Reasoning  \\
         ASDiv & 2249 & 38.2 & Mathematical Reasoning \\
        \bottomrule
    \end{tabular}}
    \caption{Dataset statistics, where “Num.” represents the number of sampled datasets, and “Length” is the number of average tokens in the sampled dataset.}
    \vspace{-2mm}
    \label{app:tab:dataset}
\end{table}

\subsection{Baselines.}
\label{app:details:baseline}

\paragraph{Probas-mean \& Probas-min: } Based on the preliminaries introduced in Section \ref{sec: prelimi}, we exemplify the following formulation for common aggregated probabilities strategies to compute the Probas-mean baseline on the entire generated answer:
\begin{equation}
    c = \frac{1}{N} \sum_{i=1}^{N} \mathbb{P} \left( y_i \mid p, y_1, \dots, y_{i-1} \right)
\end{equation}
Probas-min can be formalized as follows,
\begin{equation}
    c = \min_{i \in \{1, \dots, N\}} \mathbb{P} \left( y_i \mid p, y_1, \dots, y_{i-1} \right)
\end{equation}

\paragraph{{\small TOKEN}\textit{SAR}: }
Token-Level Shifting Attention to Relevance (\textbf{{\small TOKEN}\textit{SAR}}) is a component of the complete \textit{SAR} method \citep{duan2024shifting} that corrects generative inequalities by reviewing the relevance of each token and emphasizing uncertainty quantification attention to those more relevant components. Formally, given a sentence $s_j$ regarding prompt $x$ with the normalized relevance score for each token $z_i$ termed as $\tilde{R}_T(z_i, s_j, x)$,  the uncertainty proportions of relevant tokens are enlarged by re-weighting token entropy according to their respective $\tilde{R}_T$:
\begin{equation}
    E_T(z_i, s_j, x) = -\log p(z_i \mid s_{<i}, x) \tilde{R}_T(z_i, s_j, x).
\end{equation}
The token-level shifted ({\small TOKEN}\textit{SAR}) predictive entropy defined over $s_j$ can be formulated as,
\begin{equation}
    \text{{\small TOKEN}\textit{SAR}}(s_j, x) = \sum_{i}^{N_j} E_T(z_i, s_j, x).
\end{equation}

\paragraph{P(True): } 
We follow \citet{kadavath2022language} and prompt the LLM to judge whether its answer is correct. Our prompt followed the following template:

\begin{tcolorbox}[
    colframe=black,  
    colback=white,   
    rounded corners=north, 
    arc=1mm, 
    boxrule=.4mm, 
    width=\linewidth, 
    coltitle=black, 
    colbacktitle=white,
    title={\textbf{P(True)}}, 
    fonttitle=\bfseries, 
]

\textbf{Question}: [Question $q$]

\textbf{A student submitted}: [LLM Answer]

\vspace{5pt}

Is the student’s answer:\\
(A) True\\
(B) False

\vspace{5pt}

\textbf{The student's answer is}: 

\end{tcolorbox}

\paragraph{Self-Probing: } 
We follow \citet{xiong2023can} and prompt the LLM with a question and its answer, then asked, \textit{"How likely is the above answer to be correct"?} The procedure involves generating the answer in one chat session and obtaining its verbalized confidence in another independent chat session. Our prompt followed the following template:

\begin{tcolorbox}[
    colframe=black,  
    colback=white,   
    rounded corners=north, 
    arc=1mm, 
    boxrule=.4mm, 
    width=\linewidth, 
    coltitle=black, 
    colbacktitle=white,
    title={\textbf{Self-Probing}}, 
    fonttitle=\bfseries, 
]
\textbf{Question}: [Question $q$]

\textbf{Possible answer}: [LLM Answer]

\vspace{5pt}

\textbf{Q}: How likely is the above answer to be correct? Please first show your reasoning concisely and then answer with the following format:

\textbf{Confidence}: [the probability of answer [LLM Answer] to be correct, not the one you think correct, please only include the numerical number]\%
\end{tcolorbox}

\section{Implementation Details}
\label{app:imple}

\subsection{LLM Hyperparameters.}
\label{app:imple:hyper}
For all LLMs, the max length of each generation is set to 128 tokens. The temperature of generation is respectively set to 1.0 for Llama 3-8B and 1.2 for Llama 2-13B, and other hyperparameters as default. Besides, the hyperparameters during the P(True) process are set following \citet{kadavath2022language}, where the max token length is 1 and the temperature is followed as above setting.
\subsection{Prompts.}
\label{app:imple:prompt}

\paragraph{Prompts in Stage 1 of CoT-UQ.}
As illustrated in Section \ref{sec: method: stage1} of the main text, Stage 1 of CoT-UQ contains three separate steps: Reasoning Extraction, Keywords Extraction, and Importance Scoring. To minimize the forward times of the LLM for the sake of computational efficiency, we merge Keyword Extraction and Importance Scoring into the same prompt under a one-shot setting. Specifically, the prompts for the first three steps are as follows,

\begin{tcolorbox}[
    colframe=black,  
    colback=white,   
    rounded corners=north, 
    arc=1mm, 
    boxrule=.4mm, 
    width=\linewidth, 
    coltitle=black, 
    colbacktitle=white,
    title={\textbf{Step 1: Reasoning Extraction}}, 
    fonttitle=\bfseries, 
]

Please reason the following question step by step. Label each reasoning step as \textit{"Step i:"}, where \textit{"i"} is the step number.

You need to ensure that each step builds on the previous one and contributes meaningfully toward reaching the final answer.

Once you finish all steps, put your final answer on a separate line after the reasoning steps, starting with \textit{"Final Answer:"} (do not label it as a step).

\vspace{5pt}

\textbf{Question}: [Question $q$]  

\textbf{Response}: Let's think step by step.

\end{tcolorbox}

\begin{tcolorbox}[
    colframe=black,  
    colback=white,   
    rounded corners=north, 
    arc=1mm, 
    boxrule=.4mm, 
    width=\linewidth, 
    coltitle=black, 
    colbacktitle=white,
    title={\textbf{Step 2 \& Step 3: Keywords Extractions and Importance Scoring}}, 
    fonttitle=\bfseries, 
]

You will be provided with a question and a multi-step response containing reasoning steps. 

For each long reasoning step labeled \textit{"Step i:"}, extract the keywords, only the relevant tokens for that specific reasoning step.

You also need to evaluate the importance of each keyword to the final answer. Please evaluate the importance score following with the keyword by (/<importance score>/) on a scale of 1 to 10, where 1 is the least critical and 10 is the most critical.

If you find more than one keyword in a specific step, separate them with “;”.

If a specific step does not contribute meaningfully to deriving the final answer (e.g., repeating information already provided in the question, introducing irrelevant assumptions or speculations), return \textit{"Step i: NO ANSWER"} for that step. For example:

\vspace{5pt}

\textbf{Q}: [Question $q$]  

\textbf{A}: [Multi-Step Response $s_{1 \sim k}$]

\textbf{Keywords for Each Reasoning Step}: [Extracted Keywords and Corresponding Importance $\mathcal{K}$]

\vspace{5pt}
The following is your task:

\textbf{Q}: [Question $q$]  

\textbf{A}: [Multi-Step Response $s_{1 \sim k}$]

\textbf{Keywords for Each Reasoning Step}:
\end{tcolorbox}

It is worth noting that if the number of extracted keywords $n_i=0$ for the $i$-th step, we will label that step as \textit{"Step i: NO ANSWER"}, indicating it does not contribute meaningfully to deriving the final answer.
We provide the example introduced in the prompt template for each dataset in Table \ref{app:tab:example}.

\paragraph{Prompts in Stage 2 of CoT-UQ.}
As illustrated in Section \ref{sec: method: stage2} of the body of the paper, the Stage 2 of CoT-UQ is responsible for integrating the extracted information from reasoning path into the UQ process, where only the self-evaluation (\textit{SE}) strategy need to re-prompt the LLM.
For the variations on prompts, we present them based on the four strategies proposed in the \textit{SE} phase, namely, {\small ALL}\textit{Steps}, {\small KEY}\textit{Step}, {\small ALL}\textit{Keywords}, and {\small KEY}\textit{Keywords}. The following is the refined prompt template $p_r$ used by CoT-UQ in the \textit{SE} strategy.

\begin{itemize}
    \item \textbf{{\small ALL}\textit{Steps}}:

    \begin{tcolorbox}[
        colframe=black,  
        colback=white,   
        rounded corners=north, 
        arc=1mm, 
        boxrule=.4mm, 
        width=\linewidth, 
        coltitle=black, 
        colbacktitle=white,
        title={\textbf{P(True) \textcolor{blue}{w/ {\small ALL}\textit{Steps}}}}, 
        fonttitle=\bfseries, 
    ]
    
    \textbf{Question}: [Question $q$]
    
    \textbf{A student submitted}: [LLM Answer $a$]

    \textcolor{blue}{\textbf{The student explained the answer, which included a step-by-step reasoning: } [Multi-Step Response $s_{1 \sim k}$]}
    
    \vspace{5pt}
    
    \textcolor{blue}{Considering these reasoning steps as additional information}, is the student’s answer:\\
    (A) True\\
    (B) False
    
    \vspace{5pt}
    
    \textbf{The student's answer is}: 
    
    \end{tcolorbox}

    \begin{tcolorbox}[
        colframe=black,  
        colback=white,   
        rounded corners=north, 
        arc=1mm, 
        boxrule=.4mm, 
        width=\linewidth, 
        coltitle=black, 
        colbacktitle=white,
        title={\textbf{Self-Probing \textcolor{blue}{w/ {\small ALL}\textit{Steps}}}}, 
        fonttitle=\bfseries, 
    ]
    \textbf{Question}: [Question $q$]
    
    \textbf{Possible answer}: [LLM Answer $a$]

    \textcolor{blue}{\textbf{A step-by-step reasoning to the possible answer:} [Multi-Step Response $s_{1 \sim k}$]}
    
    \vspace{5pt}

    \textbf{Q}: \textcolor{blue}{Considering these reasoning steps as additional information}, how likely is the above answer to be correct? Please first show your reasoning concisely and then answer with the following format:
    
    \textbf{Confidence}: [the probability of answer [LLM Answer] to be correct, not the one you think correct, please only include the numerical number]\%
    \end{tcolorbox}
    
    \item \textbf{{\small KEY}\textit{Step}}:

    \begin{tcolorbox}[
        colframe=black,  
        colback=white,   
        rounded corners=north, 
        arc=1mm, 
        boxrule=.4mm, 
        width=\linewidth, 
        coltitle=black, 
        colbacktitle=white,
        title={\textbf{P(True) \textcolor{blue}{w/ {\small KEY}\textit{Step}}}}, 
        fonttitle=\bfseries, 
    ]
    
    \textbf{Question}: [Question $q$]
    
    \textbf{A student submitted}: [LLM Answer $a$]

    \textcolor{blue}{\textbf{The student explained the answer, where the most critical step is: } [Key Step $s^*$]}
    
    \vspace{5pt}
    
    \textcolor{blue}{Considering this critical reasoning step as additional information}, is the student’s answer:\\
    (A) True\\
    (B) False
    
    \vspace{5pt}
    
    \textbf{The student's answer is}: 
    
    \end{tcolorbox}

    \begin{tcolorbox}[
        colframe=black,  
        colback=white,   
        rounded corners=north, 
        arc=1mm, 
        boxrule=.4mm, 
        width=\linewidth, 
        coltitle=black, 
        colbacktitle=white,
        title={\textbf{Self-Probing \textcolor{blue}{w/ {\small KEY}\textit{Step}}}}, 
        fonttitle=\bfseries, 
    ]
    \textbf{Question}: [Question $q$]
    
    \textbf{Possible answer}: [LLM Answer $a$]

    \textcolor{blue}{\textbf{The most critical step in reasoning to the possible answer:} [Key Step $s^*$]}
    
    \vspace{5pt}

    \textbf{Q}: \textcolor{blue}{Considering this critical reasoning step as additional information}, how likely is the above answer to be correct? Please first show your reasoning concisely and then answer with the following format:
    
    \textbf{Confidence}: [the probability of answer [LLM Answer] to be correct, not the one you think correct, please only include the numerical number]\%
    \end{tcolorbox}

    \item \textbf{{\small ALL}\textit{Keywords} \& {\small KEY}\textit{Keywords}}: Prompts for keywords share the same template, except for the specific keywords content.

    \begin{tcolorbox}[
        colframe=black,  
        colback=white,   
        rounded corners=north, 
        arc=1mm, 
        boxrule=.4mm, 
        width=\linewidth, 
        coltitle=black, 
        colbacktitle=white,
        title={\textbf{P(True) \textcolor{blue}{w/ \textit{Keywords}}}}, 
        fonttitle=\bfseries, 
    ]
    
    \textbf{Question}: [Question $q$]
    
    \textbf{A student submitted}: [LLM Answer $a$]

    \textcolor{blue}{\textbf{The student explained the answer, which included the following keywords: } [Keywords set $\mathcal{K}/\mathcal{K}^*$]}
    
    \vspace{5pt}
    
    \textcolor{blue}{Considering these keywords as additional information}, is the student’s answer:\\
    (A) True\\
    (B) False
    
    \vspace{5pt}
    
    \textbf{The student's answer is}: 
    
    \end{tcolorbox}

    \begin{tcolorbox}[
        colframe=black,  
        colback=white,   
        rounded corners=north, 
        arc=1mm, 
        boxrule=.4mm, 
        width=\linewidth, 
        coltitle=black, 
        colbacktitle=white,
        title={\textbf{Self-Probing \textcolor{blue}{w/ \textit{Keywords}}}}, 
        fonttitle=\bfseries, 
    ]
    \textbf{Question}: [Question $q$]
    
    \textbf{Possible answer}: [LLM Answer $a$]

    \textcolor{blue}{\textbf{Keywords during reasoning to the possible answer:} [Keywords set $\mathcal{K}/\mathcal{K}^*$]}
    
    \vspace{5pt}

    \textbf{Q}: \textcolor{blue}{Considering these keywords as additional information}, how likely is the above answer to be correct? Please first show your reasoning concisely and then answer with the following format:
    
    \textbf{Confidence}: [the probability of answer [LLM Answer] to be correct, not the one you think correct, please only include the numerical number]\%
    \end{tcolorbox}
    
\end{itemize}

\subsection{Computational Costs Analysis}
CoT-UQ is more generation-efficient compared to previous methods based on response sampling. We have counted the time consumed for each step in the overall uncertainty
quantification pipeline, which takes about 12 seconds per sample and a total of approximately 50 GPU hours to derive all reported results for 14653 samples.
All the experiments are conducted on a server with an Intel(R) Xeon(R) Gold 5218R CPU and 8 NVIDIA A6000 GPUs.

Additionally, we further analyze the actual computational overhead associated with CoT-UQ compared to the baseline methods. The baselines in our experiments are essentially not “single-pass”. They inherently involve two inference steps: answer generation and uncertainty aggregation/self-evaluation. The additional overhead introduced by CoT-UQ only comes from reasoning chain generation and keyword extraction/scoring. Based on our empirical setting, generating an answer alone requires 3 (± 0.2) seconds per question, incorporating the CoT adds 3.5 (± 0.3) seconds, and the subsequent keyword extraction and scoring steps add 3.5 (± 0.5) seconds. Overall, CoT-UQ incurs less than 8 seconds of additional computational overhead per question. In contrast, sampling-based UQ methods generally require at least 10 answers per question to ensure robust performance \cite{qiu2024semantic}, leading to over 30 seconds per question. Thus, our method is significantly more efficient in practice.

\section{Additional Experimental Results and Further Discussion}
\label{app:add}

\subsection{Comprehensive Validation Across Model Families and Metrics}
In this part, we present experimental results to further validate the generalizability, effectiveness, and compatibility of our approach.

To validate the generalizability of our approach, we conduct experiments based on different model families since uncertainty quantification performance can vary significantly across model families. Specifically, Table \ref{app:tab:models} explicitly reports the performance of CoT-UQ based on Mistral 7B across various tasks and datasets compared to baseline methods. The results provide further evidence of the general applicability of our proposed approach.

In addtion, to comprehensively demonstrate the validity of our method, we report experimental results including the AUPRC metric in Table \ref{app:tab:metrics}. These results indicate that our CoT-UQ approach consistently outperforms baseline methods under AUPRC evaluation, further confirming its effectiveness and robustness.

Furthermore, we investigate whether incorporating CoT for fact-checking affects the overall performance of the models. As shown in Table \ref{app:tab:metrics}, our findings reveal that CoT in fact provides consistent accuracy improvements across different datasets. 
It is also important to note that all UQ methods in the paper, including CoT-UQ, do not affect the model's accuracy, as the uncertainty indicator is computed/evaluated after the model has generated its response.

\begin{table}[t]
    \centering
    \resizebox{0.99\linewidth}{!}{\begin{tabular}{l|l|cccc}
        \toprule
        \textbf{Strategy}& \textbf{Method} & \textbf{HotpotQA} & \textbf{2WikiMHQA} & \textbf{SVAMP} & \textbf{ASDiv} \\
        \midrule
         \multirow{4}*{\textit{AP}} & Probas-mean & 54.63 & 58.19 & 59.88 & 57.26 \\
         & \textbf{w/ CoT-UQ} & \textbf{58.24} & \textbf{65.58} & \textbf{65.83} & \textbf{67.40} \\
         \cmidrule{2-6}
         & {\small TOKEN}\textit{SAR} & 55.62 & 55.60 & 60.69 & 58.43 \\
         & \textbf{w/ CoT-UQ} & \textbf{58.59} & \textbf{66.06} & \textbf{65.42} & \textbf{68.18} \\
         \midrule
         \multirow{2}*{\textit{SE}} & P(True) & 48.02 & 47.13 & 42.93 & 48.44 \\
         & \textbf{w/ CoT-UQ} & \textbf{50.18} & \textbf{49.78} & \textbf{50.37} & \textbf{51.30} \\
        \bottomrule
    \end{tabular}}
    \caption{AUROC ($\uparrow$) Performance Comparison with baselines based on Mistral 7B.}
    \label{app:tab:models}
\end{table}

\begin{table}[t]
    \centering
    \resizebox{0.99\linewidth}{!}{
        \begin{tabular}{l|cccc}
            \toprule
             \multirow{2}*{\textbf{Method}} & \multicolumn{2}{c}{\textbf{HotpotQA}} & \multicolumn{2}{c}{\textbf{ASDiv}} \\ 
             \cmidrule{2-3}\cmidrule{4-5}
             & AUROC & AUPRC & AUROC & AUPRC \\
            \midrule
            Probas-mean & 53.73 & 29.14 & 58.34 & 69.03 \\
            \textbf{w/ CoT-UQ} & \textbf{62.01} & \textbf{32.25} & \textbf{64.52} & \textbf{73.95} \\
            \midrule
            {\small TOKEN}\textit{SAR} & 53.57 & 28.41 & 58.71 & 68.60 \\
            \textbf{w/ CoT-UQ} & \textbf{61.07} & \textbf{31.27} & \textbf{66.91} & \textbf{75.43} \\
            \midrule
            P(True) & 62.39 & 35.68 & 47.23 & 60.28 \\
            \textbf{w/ CoT-UQ} & \textbf{63.10} & \textbf{37.29} & \textbf{53.20} & \textbf{64.15} \\
            \bottomrule
        \end{tabular}
    }
    \caption{Comparison with baselines including AUPRC ($\uparrow$) evaluation based on Llama 3.1-8B.}
    \label{app:tab:metrics}
\end{table}

\begin{table}[t]
    \centering
    \resizebox{0.99\linewidth}{!}{
        \begin{tabular}{l|cccc}
            \toprule
             \multirow{2}*{\textbf{Datasets}} & \multicolumn{2}{c}{\textbf{w/o CoT}} & \multicolumn{2}{c}{\textbf{w/ CoT}} \\ 
             \cmidrule{2-3}\cmidrule{4-5}
             & Acc. & AUROC & Acc. & AUROC \\
            \midrule
            HotpotQA & 34.2 & 56.27 & \textbf{37.1} & \textbf{66.56} \\
            2WikiMHQA & 28.2 & 51.54 & \textbf{32.4} & \textbf{63.29} \\
            GSM8K & 17.8 & 53.96 & \textbf{18.9} & \textbf{58.54} \\
            SVAMP & 38.6 & 54.48 & \textbf{42.2} & \textbf{57.37} \\
            ASDiv & 42.4 & 57.73 & \textbf{47.8} & \textbf{59.44} \\
            \bottomrule
        \end{tabular}
    }
    \caption{Accuracy ($\uparrow$) w/ and w/o CoT based on Llama 2-13B. We report the AUROC results based on Probas-mean method.}
    \label{app:tab:metrics}
\end{table}		

\subsection{Sensitivity to Keywords Filtering threshold in \textit{SE}}
\label{app:add:importance}
\begin{figure}[t]
  \centering
  \includegraphics[width=0.48\textwidth]{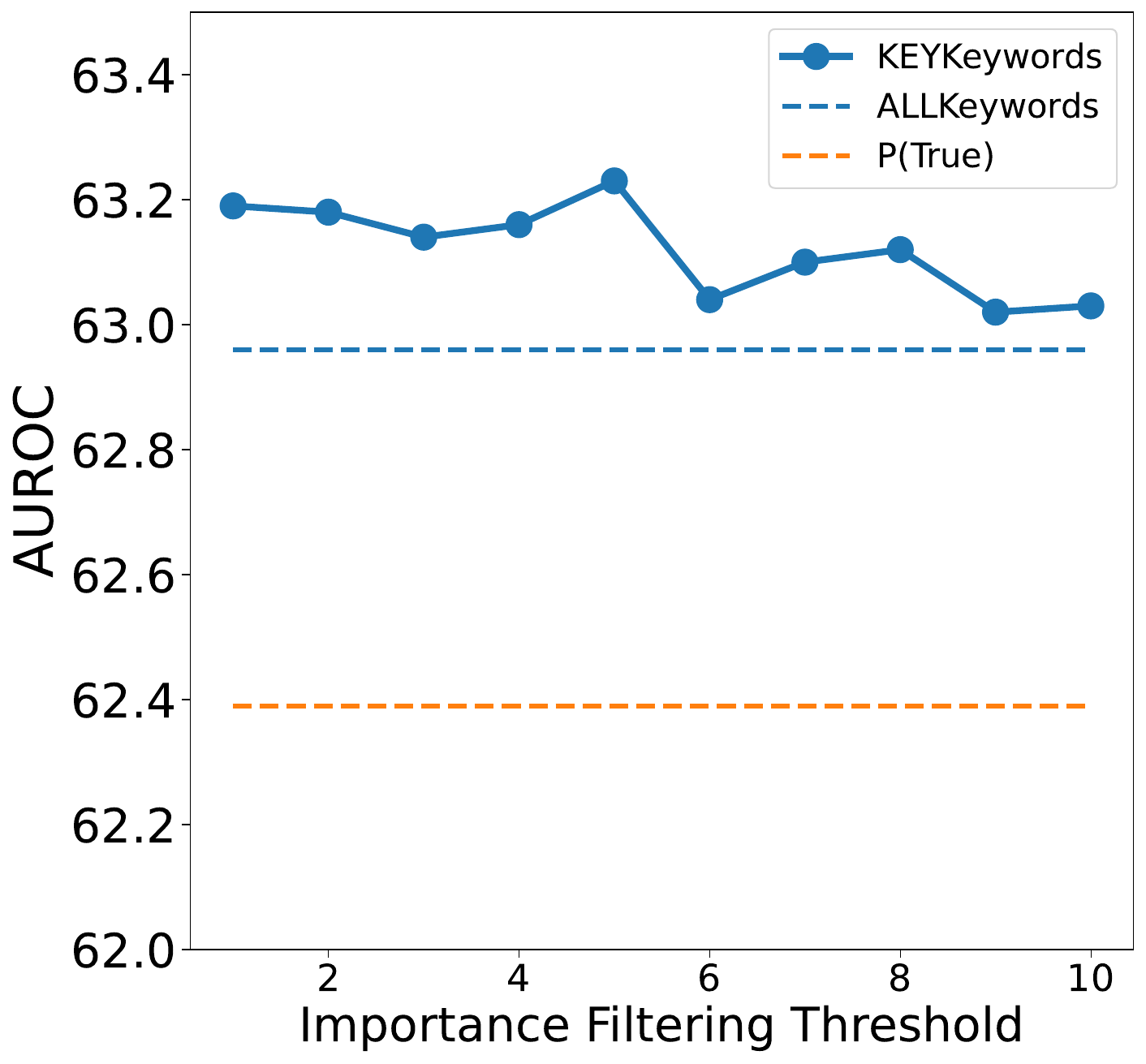}
  \vspace{-6mm}
  \caption{Analysis of the sensitivity to the importance filtering threshold $\tau$ on the HotpotQA benchmark. The experiments are based on Llama 3.1-8B.} 
  \vspace{-2mm}
  \label{app:fig:importance}
\end{figure}

To study how the {\small KEY}\textit{Keywords} is affected by the importance filtering threshold $\tau$, we conducted experiments for $\tau$ ranging from 1 to 10 on logical reasoning tasks that we have suggested using {\small KEY}\textit{Keywords} before. 
To ensure that the {\small KEY}\textit{Keywords} set $\mathcal{K}^*$ is not empty, we apply the following strategy: if the number of keywords with an importance score above the threshold $\tau$ is fewer than three, we select the top three keywords in descending order of importance as {\small KEY}\textit{Keywords} set $\mathcal{K}^*$.
Figure \ref{app:fig:importance} presents the correlations between the performance of {\small KEY}\textit{Keywords} and $\tau$. It is shown that our {\small KEY}\textit{Keywords} method is not particularly sensitive to $\tau$, but performs favorably when $\tau$ takes an intermediate value, and the results consistently outperform the {\small ALL}\textit{Keywords} and baseline P(True) methods in the logical reasoning task.

\begin{figure}[t]
  \centering
  \includegraphics[width=0.48\textwidth]{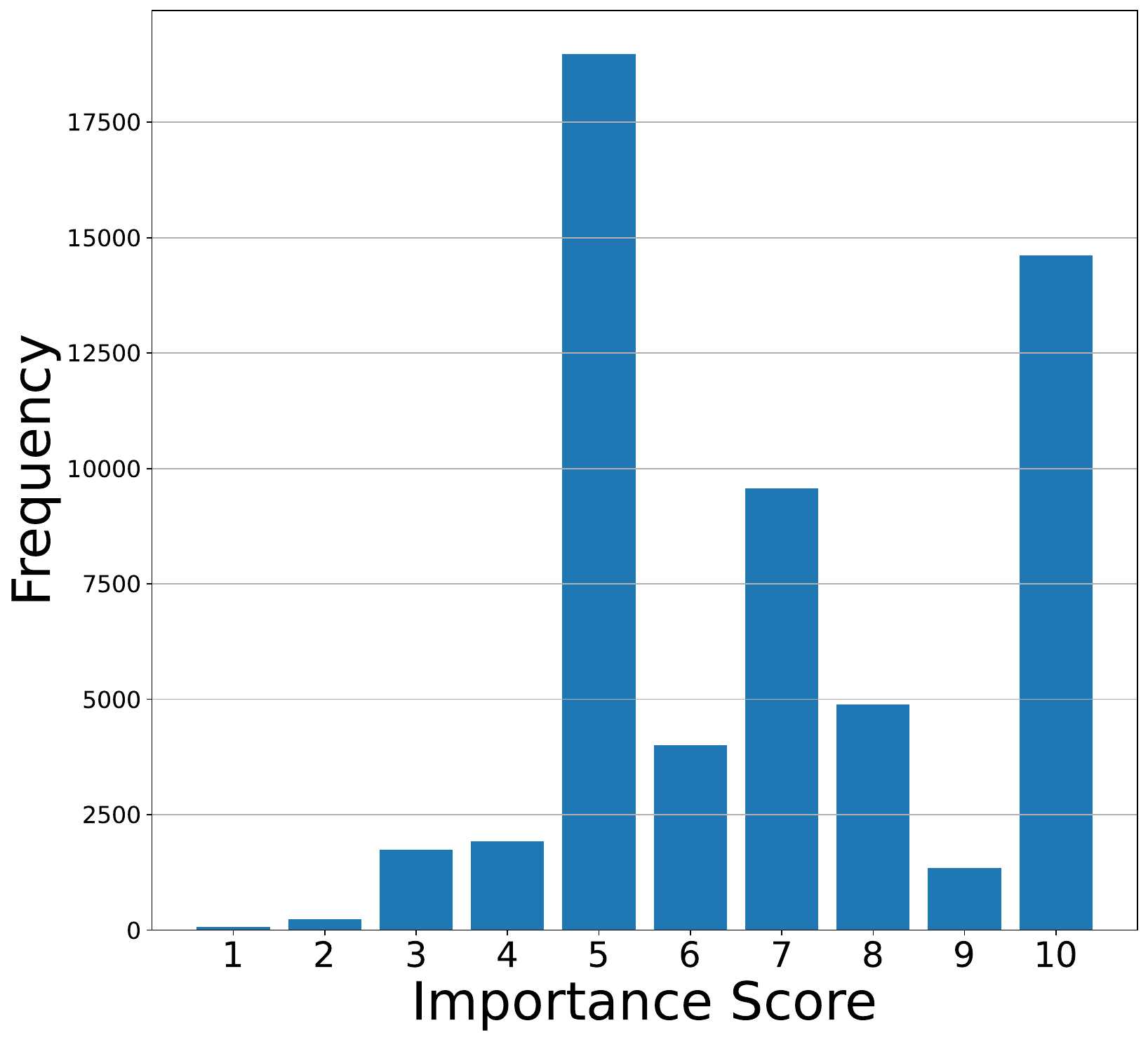}
  \caption{Frequency distribution of importance scores assigned by Llama 3.1-8B on the HotpotQA dataset.} 
  \label{app:fig:distribution}
\end{figure}

\subsection{Visualization of Model's Importance Scores}
We presented the frequency distribution of importance scores assigned by Llama 3.1-8B to keywords extracted from reasoning chains in the HotpotQA dataset, as shown in Figure \ref{app:fig:distribution}.
The results show that the LLM assigns diverse importance scores to the extracted keywords. The distribution is not concentrated on a single score but spans across the entire spectrum, indicating that the model is capable of differentiating the relative importance of different keywords. This demonstrates that LLMs are able to evaluate which parts of the reasoning process are more critical for the final answer.

\begin{table*}[t]
    \centering
    \caption{Provided Examples in \textbf{Step 2 \& Step 3} prompt template (Appendix \ref{app:imple:prompt}) for different datasets across the two reasoning domains.} 
    \label{app:tab:example}
    \begin{tabular}{@{} l p{12cm} @{}}
        \toprule
        \textbf{Dataset} & \textbf{Provided Example} \\
        \midrule
        HotpotQA & {\small \textbf{Q:} Which band has more members, "We Are the Ocean" or "The Dream Academy"? \newline
                \textbf{A:} Let’s think step by step. \newline
                Step 1: The question is asking which band has more members. \newline
                Step 2: "We Are the Ocean" has 5 members. \newline
                Step 3: "The Dream Academy" has 3 members. \newline
                Step 4: 5 is greater than 3. \newline
                Step 5: Therefore, "We Are the Ocean" has more members.\newline
                Final Answer: We Are the Ocean \newline
                \textbf{Keywords for Each Reasoning Step:} \newline
                Step 1: NO ANSWER \newline
                Step 2: We Are the Ocean(/5/); 5(/10/) \newline
                Step 3: The Dream Academy(/5/); 3(/10/) \newline
                Step 4: greater(/7/) \newline
                Step 5: We Are the Ocean(/5/)} \\ 
        \midrule
        2WikiMHQA & {\small \textbf{Q:} Who is the paternal grandmother of Joseph Ferdinand Of Bavaria? \newline
                    \textbf{A:} Let’s think step by step. \newline
                    Step 1: Joseph Ferdinand's father was Maximilian II Emanuel, Elector of Bavaria. \newline
                    Step 2: Maximilian II Emanuel was the son of Ferdinand Maria, Elector of Bavaria, and his wife, Henriette Adelaide of Savoy. \newline
                    Step 3: As the mother of Maximilian II Emanuel, Henriette Adelaide of Savoy is the paternal grandmother of Joseph Ferdinand of Bavaria. \newline
                    Final Answer: Henriette Adelaide of Savoy \newline
                    \textbf{Keywords for Each Reasoning Step:} \newline
                    Step 1: father (/8/); Maximilian II Emanuel, Elector of Bavaria (/8/) \newline
                    Step 2: son (/8/); Ferdinand Maria, Elector of Bavaria (/5/); Henriette Adelaide of Savoy (/9/) \newline
                    Step 3: mother (/10/)} \\
        \midrule
        \parbox[t]{3cm}{GSM8K \\ SVAMP \\ ASDiv} & {\small \textbf{Q:} A robe takes 2 bolts of blue fiber and half that much white fiber. How many bolts in total does it take? \newline
                \textbf{A:} Let’s think step by step. \newline
                Step 1: Identify the amount of blue fiber needed. The robe requires 2 bolts of blue fiber. \newline
                Step 2: Determine the amount of white fiber needed. It is half the amount of blue fiber, which is 2 ÷ 2 = 1 bolt. \newline  
                Step 3: Compute the total number of bolts. Add the bolts of blue fiber (2) and white fiber (1) to get 3 bolts. \newline
                Final Answer: 3 \newline
                \textbf{Keywords for Each Reasoning Step:} \newline
                Step 1: 2 bolts (/3/) \newline
                Step 2: 1 bolt (/10/) \newline
                Step 3: 3 bolts (/7/)} \\
        \bottomrule
    \end{tabular}
\end{table*}

\begin{table*}[t]
    \centering
    {\begin{tabular}{p{0.95\linewidth}}
        \toprule
        \textbf{Question}: Which involved a country farther east, In Search of the Second Amendment or Without the King? \\
        \textbf{Response}: \\
        \textit{Step 1:} In Search of the Second Amendment took place in France. \\
        \textit{Step 2:} Without the King took place in Spain.\\
        \textit{Step 3:} France is located further east than Spain.\\
        \textit{Step 4:} Therefore, In Search of the Second Amendment involved a country farther east. \\
        \textit{Final Answer:} \textcolor{red}{In Search of the Second Amendment} \\
        \textcolor{darkgreen}{\textbf{Correct Answer}: Without the King} \\
        \midrule
        \textbf{Common Probas-mean:} In Search of the Second Amendment \\
        \quad \quad \quad \quad \quad \quad \quad \quad \quad \quad \quad \quad \quad \quad \quad \quad \ \ \textcolor{red}{\textbf{1.000}} \\
        \midrule
        \textbf{Keywords for Each Reasoning Step}: \\
        \textit{Step 1:} In Search of the Second Amendment(/5/); France(/10/) \\
        \textit{Step 2:} Without the King(/5/); Spain(/10/) \\
        \textit{Step 3:} located further east(/7/) \\
        \textit{Step 4:} In Search of the Second Amendment(/5/) \\
        \textbf{Keywords Probabilities based on Probas-mean}: \\
        In Search of the Second Amendment \quad France \quad Without the King \quad Spain \quad located further east \\
        \quad \quad \quad \quad \quad \quad \ \ 0.955 \quad \quad \quad \quad \quad \quad \ \ \ \ \ \textcolor{darkgreen}{\textbf{0.066}} \quad \quad \quad \ \ \ 1.000 \quad \quad \quad \ \textcolor{darkgreen}{\textbf{0.073}} \quad \quad \quad \ \ \textcolor{darkgreen}{\textbf{0.444}} \\
        In Search of the Second Amendment\\
        \quad \quad \quad \quad \quad \quad \ \ 0.966 \\
        \textbf{Probas-mean with CoT-UQ:} \ \textcolor{darkgreen}{\textbf{0.387}} \\
        \bottomrule
    \end{tabular}}
    \caption{A case study on HotpotQA demonstrating the effectiveness of CoT in \textit{AP} strategy.}
    \label{app:tab:case0}
\end{table*}

\begin{table*}[t]
    \centering
    {\begin{tabular}{p{0.95\linewidth}}
        \toprule
        \textbf{Question}: Tim gets a promotion that offers him a 5\% raise on his \$20000 a month salary. It also gives him a bonus worth half a month's salary. How much money will he make in a year? \\
        \textbf{Response}: \textit{Step 1:} Determine the amount of Tim's new salary per month. His monthly salary goes up by 5\% (0.05 × 20000 = 1000) to 21000 per month. \\
        \textit{Step 2:} Compute the amount of Tim's bonus per month. The bonus is half a month's salary, which is 10500.\\
        \textit{Step 3:} Calculate the total amount Tim earns in a year. Add the monthly salary (21000) and bonus (10500) to get 31500 per year.\\
        \textit{Final Answer:} \textcolor{red}{31500} \\
        \textcolor{darkgreen}{\textbf{Correct Answer}: 262500} \\
        \midrule
        \textbf{Common Self-Probing:} \textcolor{red}{\textbf{100\%}}.\\
        \midrule
        \textbf{Self-Probing with Reasoning Path:} \textcolor{darkgreen}{\textbf{10\%}}. \\
        \textcolor{blue}{\textbf{You are confused between "Tim's new salary per month" and "Tim's total annual salary".}} This may seem like a trivial confusion, but reasoning is about critical reasoning. In the problem, there are two important numbers: \\
        Tim's monthly salary = 20000 \\
        Tim's monthly salary after promotion = 21000 \\
        You are given 5\% more than his monthly salary, which means he will get an additional 1000 a month. \textcolor{blue}{\textbf{The problem asks about "Tim's total annual salary", not "Tim's monthly salary after promotion"}}. \\
        A possible answer is therefore 21000 × 12 + 10500 = 262500 \\
        \bottomrule
    \end{tabular}}
    \caption{A case study on GSM8K demonstrating the effectiveness of CoT in \textit{SE} strategy.}
    \label{app:tab:case1}
\end{table*}

\subsection{Case Study}
\label{app:add:case}
\paragraph{How can CoT benefit UQ in \textit{AP}?}
We first provide a case study on the HotpotQA dataset to visualize the effect of CoT-UQ on \textit{AP} strategies. Table \ref{app:tab:case0} shows an example using Probas-mean. In the standard Probas-mean method, the model assigns a probability of \textbf{1.0} to the incorrect answer, leading to an uncalibrated and misleading over-confidence score. However, by incorporating keyword-level probability adjustments, our approach assigns more nuanced confidence scores to key reasoning components (e.g., "France: 0.066", "Spain: 0.073", "located further east: 0.444"). This recalibration mitigates overconfidence in incorrect predictions and ensures a more reliable confidence estimation with a confidence of \textbf{0.387}, demonstrating the effectiveness of our method in refining uncertainty quantification through reasoning-aware adjustments.
\paragraph{How can CoT benefit UQ in \textit{SE}?}
In Section \ref{sec:exp:main}, we noted that CoT-UQ applied to Self-Evaluation (SE) strategies achieves a greater performance improvement in mathematical reasoning tasks. We also hypothesized that this improvement is due to the model's ability to identify critical errors in its original thought process. Here, we provide detailed evidence supporting this assumption. First, we start the analysis with the following question:

\textit{Why is access to the reasoning path beneficial for UQ in self-evaluation, especially for mathematical problems?}

To answer this question, we investigate the Self-Probing strategy, which self-evaluates the credibility of an LLM’s generated answers. As demonstrated in a case from GSM8K (Table \ref{app:tab:case1}), given a math question, the LLM initially generates a multi-step response that leads to an incorrect final answer. When applying the standard Self-Probing strategy, which assesses only the correctness of the final answer, the model exhibits overconfidence, assigning an 80\% certainty to its response.

However, when Self-Probing incorporates the reasoning path, it successfully calibrates its confidence to 10\%, as it identifies problematic areas in the original thought process (highlighted in blue in Table \ref{app:tab:case1}). In this case, the incorrect response misinterpreted the problem by confusing \textit{monthly salary after promotion} with \textit{total annual salary}. By accessing its reasoning, the model correctly identifies this mistake, clarifying the key misunderstanding.

This step-by-step breakdown helps pinpoint the exact logical misstep, making it easier to adjust confidence accordingly. Furthermore, access to the reasoning path allows the LLM to distinguish between \emph{calculation errors} and \emph{conceptual errors}. If the mistake were a simple arithmetic error, the model’s confidence might remain relatively high, as the reasoning itself would still be sound. However, in this case, the mistake is conceptual, requiring significantly adjusting the confidence.

Thus, incorporating the reasoning path into uncertainty quantification leads to more precise confidence calibration, enabling the model to differentiate between minor computational mistakes and fundamental conceptual misunderstandings.


\paragraph{A complete procedure of CoT-UQ. }
To clarify each component of our approach, we present a case from the HotpotQA dataset in Table \ref{app:tab:case2}, illustrating the complete CoT-UQ process.

\begin{table*}[t]
    \centering
    {\begin{tabular}{p{0.95\linewidth}}
        \toprule
        {\fontsize{10pt}{12pt}\selectfont \textbf{Question}: Which of these board games was released earlier: Lord of the Rings or Finance?} \\
        {\fontsize{10pt}{12pt}\selectfont \textbf{Correct Answer}: Finance} \\
        \\
        {\fontsize{10pt}{12pt}\selectfont \textbf{Reasoning Extraction:}} \\
        {\fontsize{10pt}{12pt}\selectfont Please reason the following question step by step. Label each reasoning step as \textit{"Step i:"}, where \textit{"i"} is the step number.} \\
        {\fontsize{10pt}{12pt}\selectfont You need to ensure that each step builds on the previous one and contributes meaningfully toward reaching the final answer.} \\
        {\fontsize{10pt}{12pt}\selectfont Once you finish all steps, put your final answer on a separate line after the reasoning steps, starting with \textit{"Final Answer:"} (do not label it as a step)} \\
        {\fontsize{10pt}{12pt}\selectfont \textbf{Response}: Step 1: Lord of the Rings was released in 2017.} \\
        {\fontsize{10pt}{12pt}\selectfont Step 2: Finance was released in 2018.} \\
        {\fontsize{10pt}{12pt}\selectfont Step 3: 2017 is earlier than 2018.} \\
        {\fontsize{10pt}{12pt}\selectfont Step 4: Therefore, Lord of the Rings was released earlier.} \\
        {\fontsize{10pt}{12pt}\selectfont Final Answer: Lord of the Rings} \\
        \\
        {\fontsize{10pt}{12pt}\selectfont \textbf{Keywords Extractions and Importance Scoring:}} \\
        {\fontsize{10pt}{12pt}\selectfont You will be provided with a question and a multi-step response containing reasoning steps.}\\ 
        {\fontsize{10pt}{12pt}\selectfont For each long reasoning step labeled \textit{"Step i:"}, extract the keywords, only the relevant tokens for that specific reasoning step.}\\
        {\fontsize{10pt}{12pt}\selectfont You also need to evaluate the importance of each keyword to the final answer. Please evaluate the importance score following with the keyword by (/<importance score>/) on a scale of 1 to 10, where 1 is the least critical and 10 is the most critical.}\\
        {\fontsize{10pt}{12pt}\selectfont If you find more than one keyword in a specific step, separate them with “;”.}\\
        {\fontsize{10pt}{12pt}\selectfont If a specific step does not contribute meaningfully to deriving the final answer (e.g., repeating information already provided in the question, introducing irrelevant assumptions or speculations), return \textit{"Step i: NO ANSWER"} for that step.}\\
        {\fontsize{10pt}{12pt}\selectfont \textbf{Question}: Which of these board games was released earlier: Lord of the Rings or Finance?} \\
        {\fontsize{10pt}{12pt}\selectfont \textbf{Multi-Step Response}: Step 1: Lord of the Rings was released in 2017.} \\
        {\fontsize{10pt}{12pt}\selectfont Step 2: Finance was released in 2018.} \\
        {\fontsize{10pt}{12pt}\selectfont Step 3: 2017 is earlier than 2018.} \\
        {\fontsize{10pt}{12pt}\selectfont Step 4: Therefore, Lord of the Rings was released earlier.} \\
        {\fontsize{10pt}{12pt}\selectfont Final Answer: Lord of the Rings} \\
        {\fontsize{10pt}{12pt}\selectfont \textbf{Keywords for Each Reasoning Step}: Step 1: Lord of the Rings(/7/); 2017(/10/)} \\
        {\fontsize{10pt}{12pt}\selectfont Step 2: Finance(/5/); 2018(/10/)} \\
        {\fontsize{10pt}{12pt}\selectfont Step 3: earlier(/9/)} \\
        {\fontsize{10pt}{12pt}\selectfont Step 4: Lord of the Rings(/8/)} \\
        \\
        {\fontsize{10pt}{12pt}\selectfont \textbf{Reasoning Enhanced Uncertainty Quantification Strategy (Exemplified by P(True) w/ {\small KEY}\textit{Keywords}):}} \\
        {\fontsize{10pt}{12pt}\selectfont \textbf{Question}: Which of these board games was released earlier: Lord of the Rings or Finance?} \\
        {\fontsize{10pt}{12pt}\selectfont \textbf{A student submitted}: Lord of the Rings} \\
        {\fontsize{10pt}{12pt}\selectfont The student explained the answer, which included the following keywords: [2017, 2018, earlier]} \\   
        {\fontsize{10pt}{12pt}\selectfont Considering these keywords as additional information, is the student’s answer:}\\
        {\fontsize{10pt}{12pt}\selectfont (A) True} \\
        {\fontsize{10pt}{12pt}\selectfont (B) False} \\
        {\fontsize{10pt}{12pt}\selectfont \textbf{The student's answer is}:} \\
        \bottomrule
    \end{tabular}}
    \caption{A case on HotpotQA demonstrates the complete procedure of CoT-UQ.}
    \label{app:tab:case2}
\end{table*}

\end{document}